\documentclass[10pt,twocolumn,letterpaper]{article}

\usepackage[pagenumbers]{wacv} 

\usepackage{graphicx}
\usepackage{amsmath}
\usepackage{amssymb}
\usepackage{booktabs}
\usepackage{enumitem}
\usepackage{mathtools}

\usepackage{multirow}
\usepackage{xcolor}
\usepackage{colortbl}
\usepackage{makecell}
\usepackage{soul}
\usepackage{algorithm}
\usepackage{algcompatible}

\DeclareMathOperator{\mds}{calculate\_mds}
\DeclareMathOperator{\loss}{loss}

\usepackage[pagebackref,breaklinks,colorlinks]{hyperref}

\usepackage[capitalize]{cleveref}
\crefname{section}{Sec.}{Secs.}
\Crefname{section}{Section}{Sections}
\Crefname{table}{Table}{Tables}
\crefname{table}{Tab.}{Tabs.}

\def\wacvPaperID{2632} 
\def\confName{WACV}
\def\confYear{2025}

\begin{document}

\title{Identify Backdoored Model in Federated Learning via Individual Unlearning}

\author{Jiahao Xu \hspace{2em} Zikai Zhang \hspace{2em} Rui Hu\\
University of Nevada, Reno \\
\texttt{\{jiahaox,zikaiz,ruihu\}@unr.edu}
}

\maketitle

\begin{abstract}
Backdoor attacks present a significant threat to the robustness of Federated Learning (FL) due to their stealth and effectiveness. They maintain both the main task of the FL system and the backdoor task simultaneously, causing malicious models to appear statistically similar to benign ones, which enables them to evade detection by existing defense methods. We find that malicious parameters in backdoored models are inactive on the main task, resulting in a significantly large empirical loss during the machine unlearning process on clean inputs. Inspired by this, we propose MASA, a method that utilizes individual unlearning on local models to identify malicious models in FL. To improve the performance of MASA in challenging non-independent and identically distributed (non-IID) settings, we design pre-unlearning model fusion that integrates local models with knowledge learned from other datasets to mitigate the divergence in their unlearning behaviors caused by the non-IID data distributions of clients. Additionally, we propose a new anomaly detection metric with minimal hyperparameters to filter out malicious models efficiently. Extensive experiments on IID and non-IID datasets across six different attacks validate the effectiveness of MASA. To the best of our knowledge, this is the first work to leverage machine unlearning to identify malicious models in FL. Code is available at \url{https://github.com/JiiahaoXU/MASA}.
\end{abstract}


\section{Introduction} \label{sec:intro}
Federated Learning (FL)~\cite{FL_OG_shake} is an emerging paradigm for training machine learning models across multiple distributed clients while preserving their data privacy. In FL, a central server coordinates a network of clients, each owning a local dataset. During the training process, the server distributes a shared global model to each client. The clients then train this model on their local datasets and send the resulting model updates back to the server. The server aggregates these updates to refine the global model for the next round of training. FL significantly reduces privacy risks by keeping the data on the client side throughout the process. 
FL has been successfully applied in various fields such as financial analysis~\cite{fin1, fin2} and remote sensing~\cite{remotesen2, remotesense}.

However, while the distributed nature of FL enhances data security, it also introduces vulnerabilities to poisoning attacks~\cite{bulyan, liu2022evil, tian2022comprehensive}. 
For example, Byzantine attacks~\cite{lie, dnc} aim to disrupt the global model's convergence. Specifically, malicious clients intentionally alter their local model updates to differ significantly from those of benign clients, thereby distorting the convergence process. Yet, this substantial deviation between malicious and benign updates offers an opportunity for server-side detection. Recently, backdoor attacks~\cite{scaling, dba, badnet, Neurotoxin, PGD, lie}, have gained significant attention due to their stealth and practical effectiveness. Specifically, backdoor attacks aim to preserve the global model's performance on clean inputs while causing it to make incorrect predictions on inputs containing a specific pre-defined feature (i.e., trigger). Since backdoor attacks have minimal impact on the main task's accuracy, the malicious local updates closely resemble benign ones~\cite{PGD, flame}, making anomaly detection much more challenging.

One of the most common ways to defend against backdoor attacks in FL is to employ a \textit{robust aggregation rule} (AGR) on the server side to handle the received local model updates~\cite{zhang2022security}. 
Existing state-of-the-art (SOTA) AGRs can generally be classified into \textit{non-filtering-based} AGRs~\cite{rlr, lockdown, geomed, Foolsgold} and \textit{filtering-based} AGRs~\cite{krum, bulyan, mesas, mm, fltrust, deepsight}. Non-filtering-based methods aim to mitigate the harmful effects of malicious parameters in the global model. However, they often fail to fully eliminate malicious impacts during aggregation and may also degrade main task performance. In contrast, filtering-based methods focus on identifying and excluding malicious local model updates to achieve maximum robustness. They typically rely on examining statistical differences (e.g., $L_1$-norm~\cite{mm, mesas}, $L_2$-norm~\cite{mm, krum, bulyan, mesas}, and Cosine Similarity~\cite{fltrust, deepsight}) between malicious and benign updates. However, due to the dual optimization objectives of malicious clients, these statistical differences are often minimal, a phenomenon known as the poison-coupling effect~\cite{lockdown}. Furthermore, as the global model approaches convergence, the statistical differences between updates shrink further, reducing the effectiveness of filtering-based AGRs in detecting malicious updates.

Through a detailed observation of the poison-coupling effect in malicious local models, we find that backdoor parameters contribute negligibly when fed with clean inputs. This observation suggests that benign and malicious models can exhibit different behaviors during machine unlearning~\cite{bourtoule2021machine}, a process aimed at removing learned information. Specifically, when unlearning the information associated with clean data, benign and malicious models show distinct behaviors in terms of convergence speed and unlearning loss, which can serve as effective metrics for anomaly detection. Motivated by this, we propose a novel AGR called \textbf{MASA}, which leverages \underline{\textbf{M}}achine unle\textbf{\underline{A}}rning on local model\underline{\textbf{S}} individu\textbf{\underline{A}}lly with pre-unlearning model fusion to identify malicious models. In MASA, the server first reconstructs the local models using the local model updates it received.
Next, the server performs machine unlearning on each reconstructed local model and tracks its training losses during the unlearning to capture its unlearning behavior. %
Given that local models can exhibit high divergence in non-IID settings, which poses significant challenges for detecting backdoored models, MASA integrates a pre-unlearning model fusion process.
This allows each local model to incorporate parameters learned from other local datasets before unlearning, reducing inconsistencies in unlearning behavior caused by non-IID data and effectively exposing backdoored models during the unlearning process.
Finally, MASA filters out model updates with unusually large unlearning losses using a novel hyperparameter-efficient anomaly detection metric.  

In summary, our main contribution is of four folds:
\begin{itemize}[leftmargin=*]
    \item We find that to preserve the performance of the main task, malicious parameters in backdoored models are less active than benign parameters when evaluated on clean inputs. Consequently, these less active parameters lead to significantly different unlearning behavior compared to benign models. This finding offers a new perspective for designing backdoor detection methods in FL.
    \item We design a new AGR called MASA, which leverages the distinct machine unlearning dynamics between backdoored and benign local models to identify backdoored models in FL. \textbf{\textit{To the best of our knowledge, this is the first work to leverage machine unlearning for identifying backdoored models in FL}}.
    \item MASA incorporates a pre-unlearning model fusion process, which significantly reduces the divergence in local models' unlearning behavior caused by non-IID data distributions among clients, helping to expose backdoored models in non-IID settings. Moreover, MASA is equipped with a hyperparameter-efficient anomaly detection metric to identify those local models with unusual unlearning loss.
    \item We conduct extensive empirical evaluations of MASA, testing its performance on IID, extreme non-IID, and extremely high attack ratio scenarios under various SOTA backdoor attacks. Results demonstrate that MASA consistently achieves superior backdoor robustness compared to SOTA defense methods.
\end{itemize}

\section{Related Works}

\textbf{Existing backdoor attacks in FL.} 
Empirical evidence has shown that FL is susceptible to backdoor attacks due to its lack of control over the local training data of clients~\cite{scaling}. The first backdoor attack, known as {Badnet}~\cite{badnet}, adheres to the standard backdoor attack strategy by embedding the backdoor trigger in the local training data of malicious clients. Following this, numerous studies have been conducted to refine and strengthen backdoor attacks in FL~\cite{badnet, scaling, PGD, dba, Neurotoxin, f3ba, a3fl, iba, alam2022perdoor}, such as {Scaling} attack~\cite{scaling}, {Projected Gradient Descent} (PGD) attack~\cite{PGD}, {Distributed Backdoor Attack} (DBA)~\cite{dba}, {Neurotoxin} attack~\cite{Neurotoxin}, {Little is Enough} (Lie) attack~\cite{lie}, etc. Recently, trigger-optimization backdoor attacks~\cite{a3fl, f3ba, iba, lyu2023poisoning, alam2022perdoor} (as known as
{adaptive attacks} in some literature) have been studied to compromise the model with optimized triggers. 

\textbf{Defend against backdoor attacks in FL.} 
Existing defense methods generally fall into two categories: filtering-based and non-filtering-based. Filtering-based methods aim to identify and exclude malicious local models from the aggregation process~\cite{krum, bulyan, mesas, mm, fltrust, deepsight, li2024backdoorindicator}, while non-filtering-based methods seek to mitigate the impact of backdoor models on the global model~\cite{rlr, lockdown, geomed, Foolsgold, adtraining_1, xie2021crfl, FedSKU, alam2023get}. 

\textit{(1) Non-filtering-based methods:} For instance, \textit{Lockdown}~\cite{lockdown} operates under the assumption that malicious parameters of backdoored models, which are used to recognize backdoor triggers, are considered unimportant by benign clients. Consequently, it applies sparsification to prune these unimportant parameters from all the local models. \textit{FedSKU}~\cite{FedSKU} aims to recover the trigger on the server side first and then removes the knowledge of the identified triggers via machine unlearning while selectively transferring the useful knowledge into a surrogate clean model using distillation. Nevertheless, malicious parameters in backdoored models are often coupled with benign parameters, meaning they also contribute to the main task. Consequently, modifying or removing these parameters can lead to a significant loss in main task performance, rendering such methods ineffective.

\textit{(2) Filtering-based methods:} On the other hand, filtering-based methods strive for the highest backdoor robustness by identifying and filtering out malicious local models/updates. For example, \textit{Multi-Metrics}~\cite{mm} computes Manhattan distance, Euclidean distance, and Cosine Similarity for each local model update with the latest global model. Afterward, it projects the value of these metrics to its corresponding principal axis and calculates a score using the covariance matrix of these values. Those model updates with a low score will be dropped. \textit{MESAS}~\cite{mesas} assesses various metrics for each local model update including $L_1$-norm, $L_2$-norm, variance, maximum value, minimum value, and the count of weights that have increased relative to the latest global model. Using these metrics, MESAS iteratively detects and eliminates malicious updates through statistical tests and clustering. However, due to the poison-coupling effect, many of these statistical metrics (e.g., $L_1$-norm, $L_2$-norm, Cosine Similarity) are often similar for both malicious and benign models, reducing the effectiveness of existing filtering methods.

In contrast to existing defense methods, our method, MASA enjoys the following advantages: \textbf{(I)} As a filtering-based approach, MASA aims to achieve the highest backdoor robustness compared to non-filtering-based methods. \textbf{(II)} Unlike existing methods that use machine unlearning to make the global model forget the trigger through a trigger reversal process, MASA eliminates the need to recover the backdoor trigger. \textbf{(III)} Instead of relying on statistical metrics from local models, MASA leverages the intrinsic nature of malicious parameters, which exhibit reduced activity when presented with clean input and conducts individual machine unlearning on each local model to accurately and robustly expose malicious local model updates. \textbf{(IV)} Existing defense methods struggle in non-IID settings, where local models vary due to diverse local datasets. In contrast, MASA addresses this challenge using a pre-unlearning model fusion. This approach allows local models to integrate global knowledge learned from data, thereby reducing divergence in their unlearning behaviors within non-IID environments. \textbf{(V)} MASA utilizes a hyperparameter-efficient anomaly detection metric, the median deviation score, to effectively identify local model updates with abnormal unlearning behaviors.

\section{Key Motivation}

\subsection{Dual objectives of malicious clients}
In a typical FL system, a set of $ n $ clients aim to collaboratively train a shared global model $ \theta \in \mathbb{R}^d $ in an iterative manner under the coordination of a central server. 
Generally, the FL problem in a benign environment can be formulated as 
$
    \min_{\theta} (1/n)\sum_{i=1}^n F_i(\theta; D_i),
    $ 
where $F_i(\cdot)$ represents the local learning objective of client $i$. For example, for a benign client $i$ performing a multi-class classification task, its local objective can be formulated as:
\begin{equation}
    F_i(\theta; D_i) \coloneq \mathbb{E}_{(z, y)\in D_i}\mathcal{L}(\theta; z, y), \label{benign_loss}
\end{equation}
where $\mathcal{L}(\cdot)$ is the cross-entropy loss function, and $(z, y)$ is a datapoint sampled from benign dataset $D_i$. The classic method to solve the FL problem iteratively is known as FedAvg~\cite{fedavg}. Specifically, at each training round $t$, client $i\in[n]$ downloads the latest global model $\theta^{t-1}$ from the server, refines the model towards optimizing its local objective to obtain an updated local model $\theta_i^t$ and then sends its local model updates $\Delta_i^t \coloneq \theta_i^t - \theta^{t-1}$ back to the server. The server refines the global model by averaging the local updates as follows: $\theta^{t} \coloneq \theta^{t-1} + (1/n)\sum_{i=1}^n \Delta_i^t$. This process repeats until the global model converges. 

For FL under backdoor attacks, the adversary compromises or injects multiple malicious clients into the system. These malicious clients poison a portion of their local datasets to enable the injection of backdoor triggers into their local models during training. Specifically, the local dataset $D_j$ of a malicious client $j$ is divided into two subsets: $D_{j, M}$ and $D_{j, B}$. $D_{j, M}$ consists of benign data used for training on the main task with the same objective in Equation~\eqref{benign_loss}. $D_{j, B}$ contains the poisoned data, generated by stamping a trigger on each data sample and modifying the ground truth label to a target label specified by the adversary. With $D_{j, B}$, malicious clients can achieve that any input containing the trigger will be misclassified as the target label instead of its correct ground truth label.
Formally, the local learning objective of a malicious client $j$ can be formulated as follows.
\begin{equation*}
    F_j(\theta; D_j) \coloneq \underbrace{\mathbb{E}_{(z, y)\in D_{j, M}}\mathcal{L}(\theta; z, y)}_{\text{main task loss}} + 
    \underbrace{\mathbb{E}_{(\widetilde{z}, \widetilde{y})\in D_{j, B}}\mathcal{L}(\theta; \widetilde{z}, \widetilde{y})}_{\text{backdoor task loss}}, \label{malicious_loss}
\end{equation*}
where $(\widetilde{z}, \widetilde{y})$ is a datapoint sampled from the poisoned data $D_{j, B}$. These dual objectives allow malicious clients to effectively inject backdoors into their local models while preserving performance on the main task. Once the server aggregates the local model updates from all clients, these backdoors can be transferred to the global model. 

\subsection{Parameter coupling in backdoored models} 
Given that the malicious clients optimize for both the main task and the backdoor task simultaneously, their local model updates are statistically similar to those of benign ones (also known as the poison-coupling effect~\cite{lockdown}), rendering existing filtering-based methods ineffective. More precisely, the parameters of a backdoored model $\theta^*$ can be decomposed as $\theta^* = \theta_M \cup \theta_B$, where $\theta_M$ represents the benign parameters related to the main task, and $\theta_B$ represents the backdoor parameters associated with the backdoor task. This decomposition holds because of the high-level independence between the main task and the backdoor task~\cite{li2021neural, rnp}; by design, backdoor attacks should not impact the model’s performance on clean inputs~\cite{badnet}. Additionally, recent observations suggest that in backdoored models, the backdoor parameters have a negligible impact during forward propagation with clean data, preserving the performance of the main task~\cite{anp}. Note that there can be overlap between $\theta_M$ and $\theta_B$, i.e., $ \theta_M \cap \theta_B \neq \emptyset$, which means that some parameters may influence both tasks. Consequently, the coupling of parameters in malicious models poses significant challenges for mitigating/detecting backdoor attacks. For example, for non-filtering-based AGRs, if parameters critical to the main task also influence the backdoor task in the global model, removing or adjusting these parameters can significantly degrade the main task performance. For existing filtering-based AGRs, malicious models often display statistical patterns similar to benign ones, rendering traditional statistical metrics ineffective. Therefore, an effective method is required that not only identifies malicious models before aggregation but also does not rely on ineffective statistical metrics to distinguish malicious and benign models. In the following subsection, we demonstrate that machine unlearning is a desired solution that can effectively induce distinct unlearning behavior differences between malicious and benign models.

\subsection{Exposing backdoored models via unlearning}

\begin{figure}[t]
    \centering
    \includegraphics[width=0.9\linewidth]{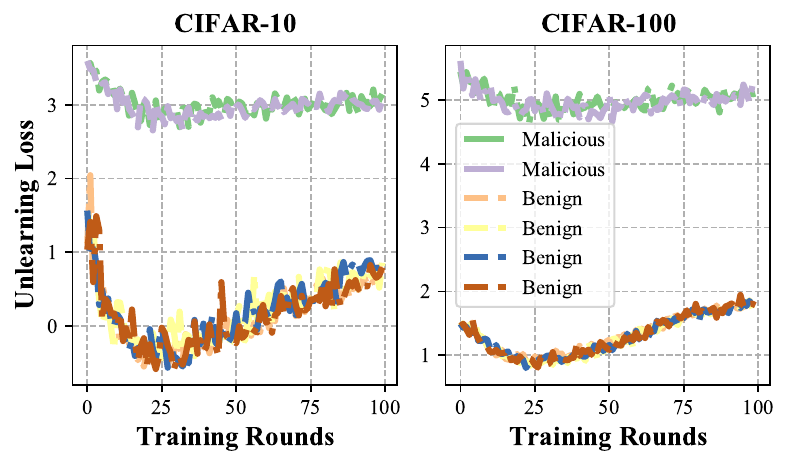}
    \vspace{-5pt}
    \caption{The unlearning loss of local models under Badnet attack on CIFAR-10 (left) and CIFAR-100 (right) w.r.t. training rounds.}
    \label{fig:intuition_loss}
    \vspace{-10pt}
\end{figure}

Since the malicious parameters $\theta_B$ are less active for the main task, one approach to identify malicious and benign models is to perform \textit{machine unlearning}~\cite{bourtoule2021machine} on all local models with respect to the main task (i.e., maximize the main task loss). Here, machine unlearning refers to a process that selectively removes specific learned information from the model. The key intuition is that machine unlearning can be regarded as a reverse process of machine learning, hence during unlearning, primarily the benign parameters $\theta_M$ are involved. As $\theta_B$ is less active in the main task, consequently, during unlearning, this reduced activity results in a \textit{faster} and \textit{more pronounced} unlearning compared to benign models. Here, the faster and more pronounced unlearning, resulting from the main engagement of $\theta_M$ in the unlearning leads to two unlearning dynamics: 1) \textit{ a larger empirical loss for backdoored models on the unlearning task}; and 2) \textit{a quicker reversion of the backdoored model to random guessing}.

To verify this intuition, we study the loss of both malicious and benign local models on unlearning the main task, and present results in \autoref{fig:intuition_loss}. Specifically, we experiment with a simple FL system where 2 of a total of 6 clients are malicious on CIFAR-10 (with ResNet18~\cite{he2016deep}) and CIFAR-100 (with VGG16~\cite{simonyan2014very}) datasets under Badnet attack~\cite{badnet}. Upon receiving the local model updates, the server reconstructs each local model and performs individual unlearning for 5 epochs.
We report the changes of averaged $\log$-transformed unlearning loss with respect to training rounds for each client. We can see that during unlearning, malicious models exhibit a significantly larger empirical unlearning loss compared to benign ones for both datasets. Additionally, benign models share similar loss patterns, as do malicious models among themselves. The distinct difference in loss patterns between benign and malicious models provides room for malicious model identification. Furthermore, the consistent loss patterns within groups of benign clients and malicious clients respectively simplify the model identification process in practice. These observations motivate the design of a novel backdoor detection method that leverages the significant differences in unlearning dynamics between malicious and benign local models. 

\section{Our Solution: MASA}
MASA is a server-side AGR that can be seamlessly integrated into existing FL systems. Specifically, upon receiving local model updates, MASA first reconstructs each local model and examines them using a well-designed individual unlearning process. Model updates exhibiting abnormal unlearning behavior are excluded from the aggregation, thereby achieving backdoor robustness of the FL system. The overall algorithm of MASA is given in Algorithm~\ref{alg: main}. 

\textbf{Individual unlearning on each local model.} The core of MASA is the \textit{individual unlearning} which unlearns each local model on the main task. To perform individual unlearning, the server must first reconstruct the local models after receiving the local model updates. Specifically, at round $t$, the server recovers the local model $\theta^t_i$ of client $i$ by
$
    \theta^t_i = \theta^{t-1} + \Delta^{t}_i,
$
where $\Delta^{t}_i$ is the local model update of client $i$ at round $t$, and $\theta^{t-1}$ is the global model at previous round. With the recovered local model, the server then conducts unlearning on a proxy clean dataset $D_p$ by solving the following minimization problem. 
\begin{equation}
    \min_{\theta^t_i} \textcolor{red}{-} \mathbb{E}_{(z, y)\in D_p}\mathcal{L}(\theta_i^t; z, y). \label{unlearning_loss}
\end{equation}
Note that the unlearning loss given in Equation~\eqref{unlearning_loss} is with a negative sign ``$\textcolor{red}{-}$" compared with the main task loss given in Equation~\eqref{benign_loss}. The server performs the unlearning process for all local models individually (line~\ref{alg_unlearning} in Algorithm~\ref{alg: main}) {and accumulates the training loss generated during unlearning 
(line~\ref{alg_calculate_loss}-~\ref{alg_accum_loss}) in order to capture the detailed unlearning behavior of a model.} Note that it is crucial for the proxy dataset to be carefully collected to ensure a certain degree of overlap with the training dataset. If this condition is not met, the unlearning process may be performed on a task unrelated to the main one, making it ineffective and potentially useless. In practice, the proxy dataset can be easily obtained by prompting cutting-edge generative models (e.g., generative adversarial networks~\cite{goodfellow2020generative} and diffusion models~\cite{ho2020denoising}).

\setlength{\textfloatsep}{10pt}
\begin{algorithm}[t]
    \caption{MASA}
    \label{alg: main}
    \begin{algorithmic}[1] 
    \REQUIRE Local updates $\{\Delta_i^t\}_{i=1}^{n}$ at round $t$, the latest global model $\theta^{t-1}$, proxy dataset $D_p$, fusion degree $\lambda$, filter radius $\delta$, unlearning epoch $E$, unlearning rate $\eta_u$.
            \STATE Initialize selection set $\mathcal{S} \leftarrow \emptyset$
            \STATE Initialize loss accumulator $\mathrm{A} \leftarrow \{\mathrm{A}_1, \mathrm{A}_2, \dots, \mathrm{A}_n \}$
            \STATE $ \bar{\Delta}^t \leftarrow (1/n) \sum_{i=1}^n \Delta_i^t$\label{alg_preaggregation}
            \FOR{$i \in [n]$}
                \STATE $\mathrm{A}_i \leftarrow 0$
                \STATE $\widetilde{\theta}_i^t \leftarrow \theta^t + \lambda \Delta_i^t + (1-\lambda) \bar{\Delta}^t \hfill\lhd$\label{alg_fusion} model fusion
                \FOR{$e=0$ to $E$}
                    \STATE $\loss \leftarrow \mathcal{L}(\widetilde{\theta_i^t}; x
)$ with each mini-batch $x$ sampled from $D_p$
                    \label{alg_calculate_loss}
                    \STATE $\mathrm{A}_i \leftarrow \mathrm{A}_i + \loss \hfill\lhd$ loss accumulation \label{alg_accum_loss}
                    \STATE $\widetilde{\theta^t_i} \leftarrow \widetilde{\theta^t_i} - \left(\textcolor{red}{-} \eta_u \nabla \mathcal{L}(\widetilde{\theta_i^t}; x) \right) \hfill\lhd$ unlearning \label{alg_unlearning}
                \ENDFOR
                
            \ENDFOR
            \FOR{client $i \in [n]$}
                \STATE $c_i \leftarrow \mds(\mathrm{A}_i, \mathrm{A}) \hfill\lhd$ by Equation~\eqref{eq: mds}\label{alg_calculate_mds}
                \IF{$c_i < \delta$} \label{alg_judge_mds}
                    \STATE $ \mathcal{S} \leftarrow \mathcal{S} \cup \{i\}$\label{alg_select} 
                \ENDIF
            \ENDFOR
            \label{alg_calculate_median}
            \STATE $ \widetilde{\Delta}\leftarrow ({1}/{|\mathcal{S}|})\sum_{i\in \mathcal{S}}{\Delta}_i^t \hfill\lhd$ benign update aggregation \label{alg_aggregate}
        \STATE \textbf{return} $\widetilde{\Delta}$
    \end{algorithmic}
\end{algorithm}

\textbf{Pre-unlearning model fusion.} In practice, clients in FL often have non-IID data, where each client only possesses a subset of the classes from the overall dataset. This leads to significant divergence among local models during training. As a result, the unlearning behavior of each model also diverges, as each model can only unlearn the information learned from samples in its local dataset. 
This variation, caused by the non-IID data, allows backdoored models to blend in with benign models, making their detection more challenging. To this end, we propose a method called \textit{pre-unlearning model fusion} to ensure the exposure of backdoored models through the  
unlearning process in non-IID settings. Specifically, before performing individual unlearning, the server constructs an aggregated model update $\bar{\Delta}$ by averaging all the received local updates (line~\ref{alg_preaggregation}). When reconstructing each local model, the server combines the local model update with the aggregated model update (line~\ref{alg_fusion}), allowing the reconstructed local model to incorporate global information. Formally, for client $i$, its local model reconstruction with pre-unlearning model fusion can be formulated as follows.
\begin{equation}
    \widetilde{\theta^t_i} = \theta^{t-1} + \lambda \Delta^{t}_i + (1-\lambda)  \Bar{\Delta}, \label{eq_fusion}
\end{equation}
where $\widetilde{\theta^t_i}$ is the reconstructed local model for client $i$, $\Bar{\Delta} := (1/n)\sum^n_{i=1} \Delta^{t}_i$, $n$ is the number of received local model updates, and $\lambda \in(0.5, 1]$ is a coefficient called \textit{fusion degree}, which controls the fusion level. In an extreme case when $\lambda=1.0$, pre-unlearning model fusion is disabled. The value of $\lambda$ much be be greater than 0.5 to ensure that the local model update predominates over the aggregated model update after fusion. With pre-unlearning model fusion, $\widetilde{\theta^t_i}$ incorporates local model updates from other clients, which contain information learned from their local datasets. This enhances the generalization ability of the reconstructed local models in non-IID settings, allowing them to perform unlearning on the main task more comprehensively and consistently, thereby reducing the divergence in unlearning behaviors across local models. 


\textbf{Efficient outlier filtering.} With the accumulated training loss values obtained in individual unlearning, MASA filters out local model updates that exhibit significantly high loss values. One approach to achieve this is by using existing clustering methods (e.g., KMeans~\cite{kmeans} and MeanShift~\cite{meanshift}) to group malicious and benign models based on their unlearning loss. However, these clustering methods often require multiple hyperparameters, necessitating time-consuming hyperparameter tuning to achieve satisfactory performance. To perform outlier filtering efficiently and accurately, we propose a new metric called \textit{Median Deviation Score} (MDS), based on the classical \textit{Z-score}. Specifically, given a set $X$ which contains $n$ elements $\{x_1, x_2, \dots, x_n \}$, the Z-score for $x_i \in X$ is calculated as $(x_i - \bar{X} )/ \sigma$ where $\bar{X}$ and $\sigma$ are the mean and the standard deviation of $X$, respectively. As for the MDS $c_i$ of $x_i$, it is calculated as 
\begin{equation}
    c_i = (x_i - \hat{X} )/ \sigma, \label{eq: mds}
\end{equation}
where $\hat{X}$ is the median element of $X$ (line~\ref{alg_calculate_mds}). Note that \textit{median} is widely used in FL as a robust metric to defend against poisoning attacks~\cite{optimalrates, signguard, LASA}. By substituting the mean with the median in the Z-score calculation, MDS becomes a more robust metric for highlighting extremely large values in a given set. Besides, we observed that the accumulated unlearning loss of malicious models is \textit{significantly} and \textit{consistently} \textbf{\textit{larger}} than that of benign ones. This allows us to utilize a single threshold parameter, called the \textit{filter radius} $\delta$, to distinguish between benign and malicious local models based on their MDS value. More precisely, local models with $c < \delta$ are considered benign and added to the selection set $\mathcal{S}$ (line~\ref{alg_judge_mds}-\ref{alg_select}). The server then aggregates the local updates in $\mathcal{S}$ to construct the global model update $\widetilde{\Delta}$ (line~\ref{alg_aggregate}), which will be used to refine the global model.

\section{Experimental Settings}


\textbf{Datasets and models.} In our experiments, we primarily use two benchmark datasets CIFAR-10~\cite{cifar10_100} and CIFAR-100~\cite{cifar10_100} and simulate both IID and non-IID data settings. For IID settings, we evenly split the dataset over the clients. We use \textit{Dirichlet distribution}~\cite{minka2000estimating} $Dir(\alpha)$ to simulate the non-IID settings with a default non-IID degree $\alpha=0.5$. We use ResNet18~\cite{he2016deep} and VGG16~\cite{simonyan2014very} as the model for CIFAR-10 and CIFAR-100, respectively. 

\textbf{Training settings.} We present the detailed training settings in Appendix Section~\ref{apdx: more_training_settings}. For all datasets, we simulate a cross-silo FL system with 20 clients. 
For the individual unlearning process of MASA, we use SGD with a learning rate of 0.001 and momentum of 0.9 for 5 epochs of training. To construct the proxy dataset on the server, we sample 1\% of the training data from the original dataset. Note that in practice the proxy dataset can be generated by cutting-edge generative models, we demonstrate that MASA with a generated proxy dataset achieves very similar performance to MASA in Appendix Section~\ref{apdx: generated_proxy_dataset}.

\textbf{Evaluation metrics.} We evaluate the performance of defense methods using three key metrics: \textit{main task accuracy (MA)}, which reflects the percentage of clean test samples that are correctly classified to their ground-truth labels by the global model; \textit{backdoor task accuracy (BA)}, which indicates the percentage of triggered samples that are misclassified to the target label by the global model; and \textit{robustness accuracy (RA)}, which measures the percentage of triggered samples that, despite the presence of the trigger, are accurately classified to their ground-truth labels by the global model. 
An effective AGR against backdoor attacks should achieve high MA and RA while maintaining a low BA.

\textbf{Evaluated attack methods.} We mainly consider six different attacks and they can be categorized by the attacker's capability to manipulate model updates into three levels: \textit{limited}, \textit{intermediate}, and \textit{advanced}. At the limited level, malicious clients are restricted to manipulating their own local datasets to generate malicious updates, which are then sent to the server for aggregation (i.e., Badnet~\cite{badnet} and DBA~\cite{dba}). At the intermediate level, attackers extend their capability by modifying the training algorithm itself in addition to dataset manipulation, resulting in more sophisticated malicious updates (i.e., Scaling~\cite{scaling} and PGD~\cite{PGD}). These two levels of attacker capabilities are commonly assumed in existing literature, where attackers control malicious devices but lack access to additional information from servers or benign clients. At the advanced level, however, attackers can access and exploit global information from the server, such as aggregated model updates, to enhance their attacks (i.e., Neurotoxin~\cite{Neurotoxin} and Lie~\cite{lie}). Note that in our work, the defense method employed by the server remains confidential to the attacker (i.e., the server is always trustable). To simulate effective backdoor attacks (achieving a BA over 60\%~\cite{mesas}), the malicious client will poison $r=50\%$ of its local data by default, where $r$ represents the \textit{data poisoning ratio}. For all the attacks, the \textit{attack ratio} is set to $f/n=20\%$ by default where $f$ is the number of malicious clients, which means 20\% of the clients in the system are malicious. Note that achieving robustness against model poisoning attacks using filtering-based methods is generally impossible when $f \geq n / 2$~\cite{liu2021approximate}. Similar to previous work on defending against backdoor attacks in FL~\cite{flame, mm}, in our work, the number of malicious clients, $f$, is assumed to be $f < n / 2$. More detailed settings of attacks are given in Appendix Section~\ref{apdx: more_attack_model_settings}.

\textbf{Evaluated defense methods.} We compare MASA with the non-robust baseline \textit{FedAvg}~\cite{fedavg} and six existing SOTA defense methods, including \textit{RLR}~\cite{rlr}, \textit{RFA}~\cite{geomed}, \textit{Multi-Krum (MKrum)}~\cite{krum}, \textit{Foolsgold}~\cite{Foolsgold}, \textit{Multi-Metric (MM)}~\cite{mm}, and \textit{Lockdown}~\cite{lockdown}. Additionally, we compare our approach to an ideal filtering-based robust aggregation, which perfectly identifies and removes all malicious updates while averaging the benign ones to update the global model. We refer to this as the most robust baseline, \textit{FedAvg*}. Note that FedAvg* achieves the highest level of robustness theoretically. 

\section{Experimental Results}

\begin{table*}[htbp]
\renewcommand{\arraystretch}{1.1}
  \centering
  \caption{The clean MA, BA, and RA of different methods on IID CIFAR-10 and CIFAR-100 under three types of attacks.}
  \scalebox{0.73}{
    \begin{tabular}{cc|c|cccc|cccc|cccc|cc}
    \toprule
    \multirow{3}[6]{*}{\textbf{\makecell*[c]{Dataset \\ (Model) }}} & \multirow{3}[6]{*}{\textbf{Method}} & \multirow{3}[6]{*}{\textbf{\makecell*[c]{Clean \\ MA$\uparrow$}}} & \multicolumn{4}{c}{\textbf{Badnet}}                    & \multicolumn{4}{c}{\textbf{DBA}}                       & \multicolumn{4}{c|}{\textbf{Scaling}} & \multirow{3}[6]{*}{\textbf{\makecell*[c]{Avg. \\ BA$\downarrow$ }}} & \multirow{3}[6]{*}{\textbf{\makecell*[c]{Avg. \\ RA$\uparrow$ }}}\\
\cmidrule(r){4-7} \cmidrule(r){8-11} \cmidrule(r){12-15}          
&   &    & \multicolumn{2}{c}{BA$\downarrow$} & \multicolumn{2}{c}{RA$\uparrow$} & \multicolumn{2}{c}{BA$\downarrow$} & \multicolumn{2}{c}{RA$\uparrow$} & \multicolumn{2}{c}{BA$\downarrow$} & \multicolumn{2}{c|}{RA$\uparrow$} \\
\cmidrule(r){4-5} \cmidrule(r){6-7} \cmidrule(r){8-9} \cmidrule(r){10-11} \cmidrule(r){12-13} \cmidrule(r){14-15}        
&    &   & $r$=0.3 & $r$=0.5 & $r$=0.3 & \multicolumn{1}{c}{$r$=0.5} & $r$=0.3 & $r$=0.5 & $r$=0.3 & \multicolumn{1}{c}{$r$=0.5} & $r$=0.3 & $r$=0.5 & $r$=0.3 & $r$=0.5 \\
    \midrule
    \multirow{9}[4]{*}{\rotatebox{90}{\makecell*[c]{CIFAR-10 \\ (ResNet18)}}} 
    & FedAvg & 91.43 & 99.98 & 99.99 & 0.02 & 0.01 & 99.93 & 99.99 & 0.07 & 0.01 & 99.98 & 99.99 & 0.02 & 0.01 & 99.98 & 0.02\\
    & FedAvg* & 91.43 & 0.56 & 0.56 & 89.03 & 89.03 & 0.56 & 0.56 & 89.03 & 89.03 & 0.56 & 0.56 & 89.03 & 89.03 & 0.56 & 89.03\\
\cmidrule{2-17}          
    & RLR   & 80.15 & 3.53 & 99.99 & 75.54 & 0.00 & \underline{0.94} & \textbf{0.72} & 78.46 & 77.12 & 1.90 & 3.00 & 80.57 & 80.10 & 18.03 & 65.63\\
    & RFA   & 89.73 & \underline{0.99} & \underline{0.90} & 87.21 & 87.43 & 1.10 & 0.99 & 87.36 & 87.78 & 1.04 & 1.06 & 87.34 & 86.17 & \underline{1.01} & 87.22\\
    & MKrum & 90.71 & 1.01 & 1.12 & \underline{88.08} & \underline{88.08} & 1.13 & 1.14 & \underline{87.64} & \underline{87.57} & \textbf{0.80} & \underline{1.01} & \underline{87.81} & \underline{87.99} & 1.03 & \underline{87.86}\\
    & Foolsgold & 91.10 & 99.97 & 99.99 & 0.03 & 0.01 & 99.92 & 99.98 & 0.07 & 0.02 & 99.99 & 99.99 & 0.01 & 0.01 & 99.97 & 0.03\\
    & MM    & 89.59 & 99.99 & 99.99 & 0.01 & 0.01 & 99.99 & 99.99 & 0.01 & 0.00 & 99.98 & 99.99 & 0.02 & 0.01 & 99.99 & 0.01\\
    & Lockdown & \textbf{91.25} & 72.71 & 46.99 & 26.33 & 49.03 & 95.59 & 29.80 & 4.34 & 57.70 & 75.22 & 10.12 & 24.13 & 74.03 & 55.74 & 39.26\\
    & \cellcolor[rgb]{ .816,  .808,  .808}MASA & \cellcolor[rgb]{ .816,  .808,  .808}\underline{91.24} & \cellcolor[rgb]{ .816,  .808,  .808}\textbf{0.90} & \cellcolor[rgb]{ .816,  .808,  .808}\textbf{0.87} & \cellcolor[rgb]{ .816,  .808,  .808}\textbf{88.30} & \cellcolor[rgb]{ .816,  .808,  .808}\textbf{88.91} & \cellcolor[rgb]{ .816,  .808,  .808}\textbf{0.81} & \cellcolor[rgb]{ .816,  .808,  .808}\underline{0.87} & \cellcolor[rgb]{ .816,  .808,  .808}\textbf{88.81} & \cellcolor[rgb]{ .816,  .808,  .808}\textbf{88.63} & \cellcolor[rgb]{ .816,  .808,  .808}\underline{0.96} & \cellcolor[rgb]{ .816,  .808,  .808}\textbf{0.87} & \cellcolor[rgb]{ .816,  .808,  .808}\textbf{88.66} & \cellcolor[rgb]{ .816,  .808,  .808}\textbf{88.91} & \cellcolor[rgb]{ .816,  .808,  .808}\textbf{0.88} & \cellcolor[rgb]{ .816,  .808,  .808}\textbf{88.70}\\
    \midrule
    \vspace{-15pt}
    &   \multicolumn{1}{c}{}    &   \multicolumn{1}{c}{}    &       &       &       & \multicolumn{1}{c}{} &       &       &       & \multicolumn{1}{c}{} &       &       &  \\
    \midrule
    \multirow{9}[4]{*}{\rotatebox{90}{\makecell*[c]{CIFAR-100 \\ (VGG16)}}} 
    & FedAvg & 67.25 & 99.65 & 99.68 & 0.26 & 0.24 & 99.64 & 99.65 & 0.32 & 0.32 & 99.83 & 99.91 & 0.11 & 0.07 & 99.73 & 0.22\\
    & FedAvg* & 67.25 & 0.83 & 0.83 & 57.09 & 57.09 & 0.83 & 0.83 & 57.09 & 57.09 & 0.83 & 0.83 & 57.09 & 57.09 & 0.83 & 57.09\\
\cmidrule{2-17}          
    & RLR   & 34.83 & 97.47 & 98.86 & 0.42 & 0.39 & 0.75 & 0.75 & 29.35 & 27.89 & 0.58 & 1.06 & 35.52 & 35.95 & 33.25 & 21.59\\
    & RFA   & 60.51 & 1.04 & \underline{0.37} & 42.25 & 44.61 & \underline{0.57} & \textbf{0.28} & 43.26 & 44.38 & 0.46 & \underline{0.39} & 47.21 & 48.95 & \underline{0.52} & 45.11\\
    & MKrum & 62.81 & \underline{0.65} & 0.60 & \underline{53.27} & \underline{53.95} & 0.70 & 0.64 & \underline{55.65} & \underline{55.46} & \underline{0.44} & 0.44 & \underline{54.05} & \underline{54.05} & 0.58 & \underline{54.41} \\
    & Foolsgold & \textbf{67.36} & 99.59 & 99.74 & 0.33 & 0.24 & 99.45 & 99.83 & 0.46 & 0.15 & 99.79 & 99.91 & 0.16 & 0.07 & 99.72 & 0.24\\
    & MM   & 65.81 & 99.92 & 99.95 & 0.07 & 0.05 & 99.96 & 99.99 & 0.03 & 0.00 & 99.79 & 99.91 & 0.17 & 0.08 & 99.92 & 0.07\\
    & Lockdown & 65.67 & 70.88 & 17.49 & 14.96 & 34.51 & 73.69 & 49.96 & 14.02 & 20.62 & 74.55 & 23.34 & 15.73 & 34.13 & 51.65 & 22.33\\
    & \cellcolor[rgb]{ .816,  .808,  .808}MASA & \cellcolor[rgb]{ .816,  .808,  .808}\underline{67.04} & \cellcolor[rgb]{ .816,  .808,  .808}\textbf{0.51} & \cellcolor[rgb]{ .816,  .808,  .808}\textbf{0.35} & \cellcolor[rgb]{ .816,  .808,  .808}\textbf{56.39} & \cellcolor[rgb]{ .816,  .808,  .808}\textbf{56.56} & \cellcolor[rgb]{ .816,  .808,  .808}\textbf{0.40} & \cellcolor[rgb]{ .816,  .808,  .808}\underline{0.48} & \cellcolor[rgb]{ .816,  .808,  .808}\textbf{56.62} & \cellcolor[rgb]{ .816,  .808,  .808}\textbf{57.93} & \cellcolor[rgb]{ .816,  .808,  .808}\textbf{0.35} & \cellcolor[rgb]{ .816,  .808,  .808}\textbf{0.35} & \cellcolor[rgb]{ .816,  .808,  .808}\textbf{56.56} & \cellcolor[rgb]{ .816,  .808,  .808}\textbf{56.56} & \cellcolor[rgb]{ .816,  .808,  .808}\textbf{0.41} & \cellcolor[rgb]{ .816,  .808,  .808}\textbf{56.77}\\
    \bottomrule
    \end{tabular}%
  }
      \vspace{-10pt}
  \label{tab: main}%
\end{table*}%

\textbf{Performance of MASA in IID settings.} We first evaluate MASA on CIFAR-10 and CIFAR-100 datasets under three attack scenarios with $r=0.3$ and $r=0.5$, respectively, and present the results for clean MA (without attacks), BA, and RA in \autoref{tab: main}. We use \textbf{bold font} to highlight the best results, while the second-best results are \underline{underlined}. Overall, MASA consistently achieves the lowest average BA and the highest average RA across both datasets, demonstrating superior backdoor robustness compared to its counterparts.

Specifically, on the CIFAR-10 dataset, RLR achieves an average BA of only 18.03\% and an RA of 65.63\%, which are 17.15\% and 23.07\% lower than those of MASA, respectively. This performance gap arises because RLR reverses the global learning direction when inconsistencies in local model parameter signs are detected. While this approach reduces some malicious influences, it degrades the global model’s performance on clean inputs and fails to fully neutralize the backdoor attack. For CIFAR-100 dataset, RFA and MKrum show slightly higher BA (0.11\% and 0.17\% higher, respectively) but considerably lower RA (11.66\% and 2.36\% lower, respectively) than MASA. RFA selects the geometric median of local model updates as the global model update, effectively preventing the global model from being compromised. However, the geometric median is suboptimal as an optimization direction compared to aggregated local model updates, leading to a lower RA due to performance loss in the main task. On the other hand, MKrum is effective in identifying malicious local model updates on IID datasets, resulting in low BA. However, MKrum's inability to select all benign local model updates (since the number of selected updates per round is typically less than the number of benign updates) leads to performance degradation in the main task, thus yielding a lower RA.

In contrast, MASA leverages the different behaviors between backdoor and benign parameters and employs individual unlearning to accurately identify and expose malicious local model updates. Moreover, MASA retains most of the benign local model updates in the aggregation process, thereby preventing performance loss in the main task.

\begin{table}[t]
  \centering
  \caption{The MA and RA results of MASA on non-IID CIFAR-100 datasets under Badnet, Lie, and Neurotoxin attacks. 
  }
  \scalebox{0.70}{
    \begin{tabular}{c|cccccc|cc}
    \toprule
    \multirow{2}[4]{*}{\textbf{Method}} & \multicolumn{2}{c}{\textbf{Badnet}} & \multicolumn{2}{c}{\textbf{Neurotoxin}} & \multicolumn{2}{c|}{\textbf{Lie}} & \multirow{2}[2]{*}{\textbf{\makecell*[c]{Avg. \\ MA$\uparrow$}}} & \multirow{2}[2]{*}{\textbf{\makecell*[c]{Avg. \\ RA$\uparrow$}}}\\
\cmidrule(r){2-3} \cmidrule(r){4-5} \cmidrule(r){6-7}          
& MA$\uparrow$   & RA$\uparrow$   & MA$\uparrow$   & RA$\uparrow$   & MA$\uparrow$  & RA$\uparrow$ \\
    \midrule
    RLR & 3.66 & 3.94 & 2.47 & 2.53 & 1.27 & 0.94 & 2.47 & 2.47 \\
    RFA & 15.11 & \underline{13.87} & 16.03 & 14.61 & 15.22 & \underline{12.83} & 15.45 & 14.57 \\
    MKrum & 49.74 & 0.42 & \underline{49.00} & 1.20 & 47.27 & 0.08 & \underline{48.67} & 16.30 \\
    Lockdown & \underline{55.71} & 6.84 & 29.42 & \underline{28.06} & \underline{55.78} & 6.06 & 46.97 & \underline{30.23} \\
    \rowcolor[rgb]{ .816,  .808,  .808} MASA  & \textbf{56.10} & \textbf{50.00} & \textbf{56.44} & \textbf{50.22} & \textbf{56.14} & \textbf{50.54} & \textbf{56.23} & \textbf{50.25} \\
    \bottomrule
    \end{tabular}%
    }
    \vspace{-5pt}
  \label{tab: non_iid}%
\end{table}%
\textbf{Performance of MASA in non-IID settings.} Here, we consider a non-IID case with $\alpha=0.1$ to simulate an extremely high data heterogeneity among clients. Such an extreme non-IID case complicates the defense against backdoor attacks in FL, as the variations in local data distributions can severely impact the consistency of local models. We report the MA and BA of MASA, RLR, MKrum, RFA, and Lockdown on CIFAR-100 datasets in \autoref{tab: non_iid}. We observed that despite the challenge of extreme non-IIDness, MASA consistently outperforms the other methods across all attack scenarios, achieving the highest MA and RA. For example, under Badnet attack, MASA attains an MA of 56.10\% and an RA of 50.00\%, outperforming other methods like Lockdown, which achieves an MA of 55.71\% and an RA of 6.84\%. Similarly, under Neurotoxin attack, MASA achieves an MA of 56.44\% and an RA of 50.22\%, significantly higher than MKrum's MA of 49.00\% and RA of 1.20\%. These results demonstrate MASA's effectiveness in maintaining both high accuracy and robustness, even under severe non-IID cases across different attacks. MASA incorporates a pre-unlearning model fusion, which effectively mitigates the challenges posed by diverse local model updates, enabling the detection of backdoored models in such intensely non-IID scenarios.

\textbf{Performance of MASA under various attack ratios.} 
\begin{figure}[t]
    \centering
    \includegraphics[width=0.9\linewidth]{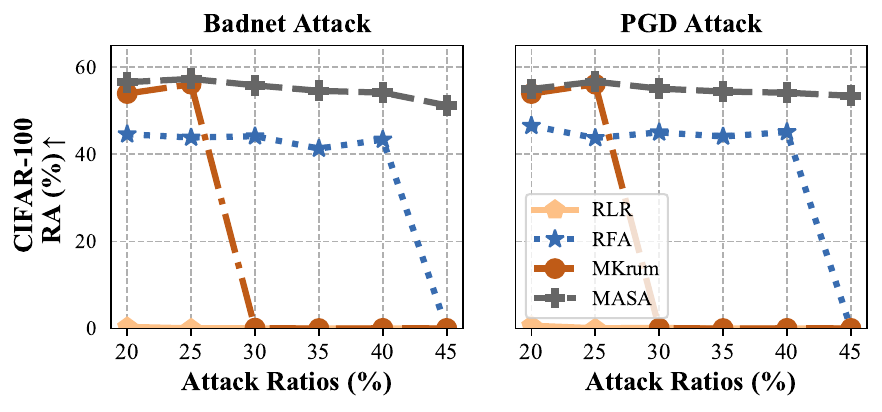}
    \vspace{-5pt}
    \caption{The RA of MASA, compared to RLR, RFA, and MKrum, under Badnet (left) and PGD (right) attacks with attack ratios ranging from 20\% to 45\%.}
    \label{fig: attackraio}
\end{figure}
We conduct experiments to validate MASA's robustness in scenarios with high attack ratios. Specifically, we vary the attack ratio from 20\% to an extreme value of 45\% for both Badnet and PGD attacks on CIFAR-100 dataset. The results, summarized in \autoref{fig: attackraio}, demonstrate that MASA consistently outperforms other methods, including RLR, RFA, and MKrum, across all attack ratios. As the attack ratio increases, MASA consistently maintains the highest robustness, though it experiences a slight drop in RA. This decrease occurs because MASA discards more local model updates in response to the growing number of malicious clients in the system. In contrast, MKrum exhibits sensitivity to attack ratios, losing its effectiveness when the attack ratio reaches 30\% or higher, and RLR fails to defend against both attacks across all attack ratios. While RFA shows some degree of robustness, it loses its effectiveness under both attacks when the attack ratio reaches 45\%.

\textbf{Effectiveness of MASA in cross-device FL with client sampling.} The majority of our experiments are conducted within the cross-silo FL setting. It is also important to consider the cross-device FL scenario, where a large number of clients participate in the system. To this end, we simulate 100 clients, with the server randomly selecting 20 clients per round to perform training. This random sampling of clients enables 1) dynamically selected local updates to contribute to refining the global model, and 2) the possibility of selecting a majority of malicious clients, creating a more dynamic and challenging environment for AGR operations.
We present the MA, BA, and RA of different baselines on CIFAR-10 under DBA and Lie attacks in \autoref{tab: cross_device}. Generally, MASA consistently achieves the best performance in terms of all evaluation metrics among the evaluated methods. Specifically, MASA achieves an average RA of 82.97\% and an average BA of 1.21\%, outperforming Mkrum with a gap of +0.37\% and +1.36\%, respectively. While Lockdown achieves competitive MA under DBA, however, it fails to provide any robustness in both cases. MASA's ability to balance high accuracy and robustness in a challenging cross-device FL environment with client sampling underscores its effectiveness and adaptability in real-world scenarios.

\begin{table}[t]
  \centering
  \caption{Performance of different methods in cross-device FL settings on CIFAR-10 dataset under DBA and Lie attacks.}
  \scalebox{0.70}{
    \begin{tabular}{c|cccccc|cc}
    \toprule
    \multirow{2}[4]{*}{\textbf{Method}} & \multicolumn{3}{c}{\textbf{DBA}} & \multicolumn{3}{c|}{\textbf{Lie}} & \multirow{2}[2]{*}{\textbf{\makecell*[c]{Avg. \\ BA$\downarrow$}}} & \multirow{2}[2]{*}{\textbf{\makecell*[c]{Avg. \\ RA$\uparrow$}}}\\
\cmidrule(r){2-4} \cmidrule(r){5-7}         & MA$\uparrow$    & BA$\downarrow$    & RA$\uparrow$    & MA$\uparrow$    & BA$\downarrow$    & RA$\uparrow$ \\
    \midrule
    FedAvg &   84.91    &   99.89    &    0.09   &   84.73    &    99.99   &  0.01 & 99.94 & 0.05\\
    FedAvg* &    85.23   &    1.43   &   81.71    &   85.23    &   1.43    &  81.71 & 1.43 & 81.71\\
    \cmidrule{1-9}
    RFA   &   83.86    &   \underline{1.71}    &    \underline{81.53}  &   \underline{84.76}    &   1.52    &  81.66 & 1.62 & 81.60\\
    Mkrum &   83.98    &   1.83    &   80.93    &   84.12    &  \underline{1.33}     & \underline{82.29} & \underline{1.58} & \underline{81.61}\\
    Lockdown &  \underline{84.14}     &    99.99   &    0.00   &    83.49   &    99.99   &  0.01 & 99.99 & 0.01\\
    \rowcolor[rgb]{ .816,  .808,  .808}MASA  &   \textbf{85.11}    &   \textbf{1.31}    &   \textbf{82.91}    &   \textbf{84.80}    &  \textbf{1.10}     & \textbf{83.03} & \textbf{1.21} & \textbf{82.97}\\
    \bottomrule
    \end{tabular}%

}
    \vspace{-6pt}
  \label{tab: cross_device}%
\end{table}%

\textbf{Hyperparameter study.} In this section, we discuss the impact of two critical hyperparameters in MASA: the fusion degree $\lambda$ and the filter radius $\delta$. The parameter $\lambda$ controls the intensity of pre-unlearning model fusion, with smaller values indicating more intense fusion, while $\delta$ determines detection sensitivity, with smaller values reflecting stricter filtering. We report the average True Positive Rate (TPR) and False Positive Rate (FPR) over training for different settings on the non-IID CIFAR-10 dataset under three different attacks in \autoref{parameter_study}. It is observed that when $\lambda=0.3$, the TPR is relatively lower compared to cases with higher $\lambda$ values. This suggests that the aggregated model update predominates in the local models, compromising their unique features excessively. Therefore, we recommend selecting the value of $\lambda$ between just above 0.5 and 1.0. It is important to note that when $\lambda=1.0$, where pre-unlearning model fusion is disabled, the FPR increases across all cases, highlighting the effectiveness of model fusion. On the other hand, $\delta$ affects the strictness of the outlier filtering process. Lower values (e.g., $\delta = 0.5$) result in significantly lower FPRs (an average of 4.50\%) but also lead to a notable reduction in TPRs (an average of 88.75\%). In contrast, higher values (e.g., $\delta = 2.0$) nearly maximize TPRs across all attacks but cause a substantial increase in FPR. The parameter $\delta$ provides flexibility in balancing TPR (main task performance) and FPR (backdoor robustness) in MASA, allowing it to be tailored to specific security requirements. In comparison, MASA with a $\lambda$ of 0.7 and a $\delta$ of 1.0 strikes an optimal balance between TPR and FPR in this setting.  We additionally conduct experiments on various proxy data sizes and present the results in Appendix Section~\ref{apdx: proxy_dataset_size}.

\begin{table}[t]
  \centering
  \caption{The TPR and FPR of MASA with different $\lambda$ and $\delta$ on non-IID CIFAR-10 datasets under Badnet, DBA, and PGD attacks.}
  \label{parameter_study}
  \scalebox{0.70}{
    \begin{tabular}{c|cccccc|cc}
    \toprule
    \multirow{2}[4]{*}{\textbf{Setting}} & \multicolumn{2}{c}{\textbf{Badnet}} & \multicolumn{2}{c}{\textbf{DBA}} & \multicolumn{2}{c|}{\textbf{PGD}} & \multirow{2}[2]{*}{\textbf{\makecell*[c]{Avg. \\ TPR$\uparrow$}}} & \multirow{2}[2]{*}{\textbf{\makecell*[c]{Avg. \\ FPR$\downarrow$}}}\\
\cmidrule(r){2-3} \cmidrule(r){4-5} \cmidrule(r){6-7}          & TPR$\uparrow$   & FPR$\downarrow$   & TPR$\uparrow$   & FPR$\downarrow$   & TPR$\uparrow$  & FPR$\downarrow$ \\
    \midrule
    $\lambda=0.3$ & 81.56 & 2.50  & 80.00 & 7.00  & 81.44 & 10.25 & 81.00 & 6.58 \\
    
    $\lambda=0.5$ & 95.13 & \underline{1.75}  & 94.75 & \underline{6.25}  & 92.81 & \underline{9.25} & 94.23 & \underline{5.75}  \\
    $\lambda=1.0$ & 98.94 & 7.00  & 98.12 & 15.50 & 97.44 & 15.00 & 98.17 & 12.50 \\
    \midrule
    $\delta=0.5$ & 89.62 & \textbf{0.25 } & 88.88 & \textbf{5.75}  & 87.75 & \textbf{7.50} & 88.75 & \textbf{4.50} \\
    $\delta=1.5$ & 97.38 & 78.00 & \textbf{99.94} & 55.50 & 97.19 & 81.50 & 98.17 & 71.67 \\
    $\delta=2.0$ & \textbf{99.88} & 94.00 & \textbf{99.94} & 93.75 & \textbf{99.19} & 98.25 & \textbf{99.67} & 95.33\\
    \midrule
    \rowcolor[rgb]{ .816,  .808,  .808} MASA  & \underline{99.25} & 2.75  & \underline{98.81} & 7.75  & \underline{97.56} & 10.00 & \underline{98.54} & 6.83 \\
    \bottomrule
    \end{tabular}%
    }
    \vspace{-6pt}
\end{table}%

\section{Conclusion}
We propose MASA to defend against stealthy backdoor attacks in FL. MASA identifies malicious parameters by leveraging individual unlearning on the main task for each local model, based on the observation that malicious parameters are less active on the main task. To address non-IID challenges, MASA incorporates a pre-unlearning model fusion mechanism, incorporating global information with local models to improve the consistency in unlearning behaviors across local models. It then filters malicious updates using a newly introduced metric, the median deviation score, based on the accumulated unlearning loss values of local models. We extensively evaluate MASA's superior performance across various scenarios and provide a detailed analysis of its hyperparameters. We further discuss the limitation of MASA and future work in Appendix Section~\ref{apdx: discussion}.
%

{\small
\bibliographystyle{ieee_fullname}
\bibliography{egbib}
}

\clearpage
\section{Appendix}
\subsection{More details of attack settings}
\label{apdx: more_attack_model_settings}
We add a ``plus'' trigger to benign samples to generate the poisoned data samples. For DBA attack~\cite{dba}, we decompose the ``plus" trigger into four local patterns, and each malicious client only uses one of these local patterns. For Scaling attack~\cite{scaling}, we use a scale factor of 2.0 to scale up all malicious model updates. For PGD attack~\cite{PGD}, malicious local models are projected onto a sphere with a radius equal to the $L_2$-norm of the global model in the current round for CIFAR-100, while for CIFAR-100 we make the radius of the sphere 10 times smaller than the norm. For Neurotoxin~\cite{Neurotoxin}, malicious model updates are projected to the dimensions that have Bottom-75\% importance in the aggregated model update from the previous round. For Lie attack~\cite{lie}, we set the maximal value $z=1.5$.

\subsection{More details of defense model}
\label{apdx: more_defense_model_settings}
In our setting, the server does not have access to the clients' local datasets but is familiar with the training objective, allowing the server to collect a proxy dataset independently which is correlated to the local data distribution. Additionally, the server lacks specific information about the backdoor attacks, such as the type of trigger used. We further assume that the server has no prior knowledge of the number of malicious clients. To defend against backdoor attacks, the server will apply an AGR to handle local model updates received from clients and generate an aggregated model update at each training round.

\subsection{More details of training settings}
\label{apdx: more_training_settings}
We use stochastic gradient descent (SGD) as the local solver, with the learning rates set as $0.1$ with the decay ratio $0.99$ and the number of local training epochs set as $2$. Note that in our setting, malicious clients share the same settings as benign ones. The number of training rounds is set to $T = 100 $ for CIFAR-10~\cite{cifar10_100} and $T = 150$ for CIFAR-100~\cite{cifar10_100}.

\subsection{Experiments on various proxy data sizes}
\label{apdx: proxy_dataset_size}
We further examine how the size of the proxy dataset affects MASA’s performance. Specifically, we vary the number of images in the proxy dataset from the default 500 (1\% of the training dataset) down to an extreme of 50 images. The MA, BA, and RA on both IID and non-IID CIFAR-10 datasets are presented in \autoref{tab: dataset_size}. For the IID case, MASA's performance remains relatively stable regardless of the proxy dataset size. Even when the proxy dataset is reduced to 50 images, MASA experiences only a slight drop in BA and RA. However, in non-IID scenarios, MASA shows greater sensitivity to proxy dataset size. MASA remains robust until the dataset size drops to 125 images, after which its ability to defend against backdoor attacks weakens significantly. Based on these results, MASA should be implemented with a reasonably sized proxy dataset in practice. Our experiments show that using a proxy dataset with a size of just 1\% of the training dataset is sufficient, which not only reduces the time and effort required for data collection but also minimizes storage needs and computational overhead. This makes MASA more practical and scalable in real-world applications where resources are limited.
\begin{table}[t]
  \centering
  \caption{The MA, BA, and RA comparison across different proxy dataset sizes.}
  \scalebox{0.7}{
    \begin{tabular}{c|c|cccccc}
    \toprule
    \multirow{2}[4]{*}{\textbf{Distribution}} & \multirow{2}[4]{*}{\textbf{Metric}} & \multicolumn{6}{c}{\textbf{Proxy dataset size $|D_p|$}}  \\ \cmidrule{3-8} &    & \cellcolor[rgb]{ .816,  .808,  .808}500   & 250  & 200  & 125   & 100   & 50 \\
    \midrule
   \multirow{3}[2]{*}{IID} 
          & MA$\uparrow$  & \cellcolor[rgb]{ .816,  .808,  .808}90.86 & 90.83 & 90.68 & 91.28 & 91.03 & 90.87 \\
   
   & BA$\downarrow$ & \cellcolor[rgb]{ .816,  .808,  .808}0.87 & 0.50 & 0.58 & 0.84 & 
 0.72 & 1.19 \\
   
   & RA$\uparrow$  & \cellcolor[rgb]{ .816,  .808,  .808}88.91 & 88.74 & 88.67 & 88.36 & 88.68 & 88.14 \\
       \midrule
    \vspace{-14pt}
          &       &       &     &   &       &       &     \\
    \midrule
       \multirow{3}[2]{*}{Non-IID} 
          & MA$\uparrow$ & \cellcolor[rgb]{ .816,  .808,  .808}88.44 & 88.05 & 88.41 & 86.46 & 87.73& 88.35 \\
   
   & BA$\downarrow$ & \cellcolor[rgb]{ .816,  .808,  .808}0.77 & 1.83 & 1.72 & 6.78 & 60.37 & 99.99\\
   
   & RA$\uparrow$ & \cellcolor[rgb]{ .816,  .808,  .808}85.21 & 84.02 & 84.21 & 77.60 & 35.58& 0.01\\
    \bottomrule
    \end{tabular}%
    }
  \label{tab: dataset_size}%
\end{table}%

\subsection{Experiments on generated proxy datasets}
\label{apdx: generated_proxy_dataset}
In our default setting, we sample 1\% of training data to construct the proxy dataset. Here, we assess MASA's performance with a proxy dataset generated by cutting-edge pre-trained generative models. Specifically, we utilize the checkpoint from the SOTA StyleGAN-XL~\cite{sauer2022stylegan}\footnote{\url{https://github.com/autonomousvision/stylegan-xl}} to generate 50 images per class of CIFAR-10 dataset, forming the proxy dataset. We refer to MASA using this generated dataset as MASA$^*$. The MA, BA, and RA under both Badnet and Scaling attacks on IID and non-IID CIFAR-10 datasets are summarized in \autoref{tab: generateddataset}. Overall, MASA$^*$ demonstrates performance consistent with MASA across both IID and non-IID scenarios. These results suggest that MASA remains effective when applied to a generated proxy dataset, significantly improving its practical utility in situations where collecting a proxy dataset is challenging or infeasible.
\begin{table}[t]
  \centering
  \caption{Performance of MASA$^*$ and MASA on IID and non-IID CIFAR-10 datasets.}
  \scalebox{0.7}{
    \begin{tabular}{c|c|cccccc}
    \toprule
    \multirow{2}[4]{*}{\textbf{Distribution}} & \multirow{2}[4]{*}{\textbf{Method}} & \multicolumn{3}{c}{\textbf{Badnet}} & \multicolumn{3}{c}{\textbf{Scaling}}  \\
\cmidrule(r){3-5} \cmidrule(r){6-8}          &       & MA$\uparrow$    & BA$\downarrow$    & RA$\uparrow$ & MA$\uparrow$    & BA$\downarrow$    & RA$\uparrow$ \\
    \midrule
    \multirow{2}[2]{*}{IID} 
          & MASA$^*$  & 90.88 & 0.71 & 88.96 & 90.88 & 0.71 & 88.96 \\
          & \cellcolor[rgb]{ .816,  .808,  .808}MASA & \cellcolor[rgb]{ .816,  .808,  .808}90.86 & \cellcolor[rgb]{ .816,  .808,  .808}0.87 & \cellcolor[rgb]{ .816,  .808,  .808}88.91 & \cellcolor[rgb]{ .816,  .808,  .808}90.86 & \cellcolor[rgb]{ .816,  .808,  .808}0.87 & \cellcolor[rgb]{ .816,  .808,  .808}88.91 \\
    \midrule
    \vspace{-14pt}
          &       &       &       &       &       &     \\
    \midrule
    \multirow{2}[2]{*}{Non-IID} 
          & MASA$^*$ & 88.52 & 1.47 & 84.31 & 88.48 & 0.80 & 85.57\\
          & \cellcolor[rgb]{ .816,  .808,  .808}MASA & \cellcolor[rgb]{ .816,  .808,  .808}88.44 & \cellcolor[rgb]{ .816,  .808,  .808}0.77 & \cellcolor[rgb]{ .816,  .808,  .808}85.21 & \cellcolor[rgb]{ .816,  .808,  .808}88.60 & \cellcolor[rgb]{ .816,  .808,  .808}0.96 & \cellcolor[rgb]{ .816,  .808,  .808}85.34 \\
    \bottomrule
    \end{tabular}%
    }
  \label{tab: generateddataset}%
\end{table}%

\subsection{Discussion and future works}
\label{apdx: discussion}
In this section, we discuss the primary limitation of MASA: the individual unlearning performed on the server adds an extra computational load. This limitation can impact MASA's effectiveness, especially in larger-scale FL deployments. One potential solution to mitigate this limitation is to utilize a more powerful server capable of parallel unlearning. This approach would reduce the computational cost of individual unlearning by a factor of $1/n$ compared to the current MASA implementation.

Another limitation of MASA is its reliance on a clean proxy dataset that overlaps with the main task data, which may conflict with the privacy-preserving goals of FL in sensitive scenarios. To address this, one possible solution is to shift the unlearning process to local execution on clients. This approach would require protection to ensure that malicious clients follow the unlearning protocol. Alternatively, a verification mechanism could be introduced to detect if the models returned by clients genuinely reflect the unlearning process, thereby maintaining robustness against malicious behavior.

\end{document}